\documentclass[letterpaper]{article} 
\usepackage{aaai25}
\usepackage{times}  
\usepackage{helvet}  
\usepackage{courier}  
\usepackage[hyphens]{url}  
\usepackage{graphicx} 
\usepackage{natbib}  
\usepackage{caption} 
\usepackage{subcaption}
\usepackage{multirow}
\usepackage{algorithm}
\usepackage{algorithmic}
\usepackage{amsmath}
\usepackage{nicefrac}

\usepackage{newfloat}
\usepackage{listings}
\DeclareCaptionStyle{ruled}{labelfont=normalfont,labelsep=colon,strut=off} 
\floatstyle{ruled}
\newfloat{listing}{tb}{lst}{}
\floatname{listing}{Listing}
\title{Bidirectional Emergent Language in Situated Environments}
\author {
    Cornelius Wolff\textsuperscript{\rm 1},
    Julius Mayer\textsuperscript{\rm 1},
    Elia Bruni\textsuperscript{\rm 1},
    Xenia Ohmer\textsuperscript{\rm 1}
}
\affiliations {
    \textsuperscript{\rm 1}Osnabrück University\\
    cowolff@uos.de, research@jmayer.ai, elia.bruni@uos.de, xenia.ohmer@uos.de
}

\usepackage{bibentry}

\begin{document}

\nocopyright
\maketitle

\begin{abstract}
Emergent language research has made significant progress in recent years, but still largely fails to explore how communication emerges in more complex and situated multi-agent systems. 
Existing setups often employ a reference game, which limits the range of language emergence phenomena that can be studied, as the game consists of a single, purely language-based interaction between the agents. 
In this paper, we address these limitations and explore the emergence and utility of token-based communication in open-ended multi-agent environments, where situated agents interact with the environment through movement and communication over multiple time-steps.
Specifically, we introduce two novel cooperative environments: Multi-Agent Pong and Collectors. 
These environments are interesting because optimal performance requires the emergence of a communication protocol, but moderate success can be achieved without one. 
By employing various methods from explainable AI research, such as saliency maps, perturbation, and diagnostic classifiers, we are able to track and interpret the agents' language channel use over time. 
We find that the emerging communication is sparse, with the agents only generating meaningful messages and acting upon incoming messages in states where they cannot succeed without coordination.
\end{abstract}

%

\section{Introduction}

Agent-based simulations of emergent communication have long been popular in evolutionary linguistics and AI research. 
More recently, starting with the work of \citet{foerster_2016} and \citet{lazaridou_2017}, there has been a growing interest in language emergence (LE) simulations involving deep neural network (DNN) agents \cite{lazaridou_2020}. 
Inspired by the Lewis Signaling Game \cite{Lewis1969}, a large portion of these DNN-based experiments use simple reference games with one sender and one receiver agent: The sender sees some target object and sends a message to the receiver, which has to identify said target object among a set of distractors. 
If the receiver is successful both agents are rewarded.
However, reference game approaches make critical simplifications, typically including uni-directional communication, full cooperation, single-message interactions, and non-situatedness.
As a consequence, they fail to capture essential aspects of communication, such as language use beyond reference, spatial and temporal dynamics, population dynamics, nonverbal communication, or deception, among others \cite[e.g.,][]{zubek2023models}. 

At the same time, multi-agent reinforcement learning (MARL) research more generally has been studying multi-agent coordination in comparatively complex scenarios, more closely tailored to capture (aspects of) real world applications. 
Given that language is a means to achieve or enhance such coordination, various studies in this context experiment with communicating agents \cite[][]{Khan_Ahmad_2023, zhu_2024}. 
In contrast to reference game setups, these approaches typically involve bidirectional communication among multiple agents, and the use of more elaborate tasks introduces aspects such as situatedness, partial observability, competition, and communication beyond reference. Thus, despite largely being motivated by practical concerns of multi-agent coordination, they are also interesting from an evolutionary linguistics perspective as they give rise to alternative, more realistic LE scenarios. 

In this work, we aim to use the types of computational experiments and training methods developed in the general context of MARL to study LE in more complex environments, thus overcoming important limitations of reference-game-based setups. In particular,
our \emph{Multi-Agent Pong} environment is inspired by the Atari Game Pong and expands the existing setup to two players, which have to prevent two balls from hitting the wall. 
In our \emph{Collectors} environment, agents have to collect as many objects as possible, with the additional challenge that objects spawn at random locations and disappear after a certain time window. 
Note that these environments distinguish themselves from reference games in important ways. 
First, the agents interact with the environment over multiple time steps within one game; and while an optimal solution of the environments requires communication, there are numerous states where communication is not important. 
Such states occur when each agent can catch both balls (Pong) or collect all targets (Collectors). 
Second, agents are situated and interact with the environment through language (as in a reference game) but also through physical movement.
As a result, agents may have moderate success without developing a communication system, they may use communication sparsely, and different communication strategies beyond naming objects can emerge, such as communicating positions or directions of movement of objects as well as agents. 

While giving cause to criticism, the simplicity of reference games also has some obvious advantages. 
Reference game simulations require comparatively little compute and simple RL algorithms such as REINFORCE \cite{williams1992simple} are sufficient for training. Furthermore, understanding whether a successful protocol has emerged and interpreting this protocol is relatively straightforward in these single-purpose, single-time-step games. Communication is successful when rewards increase and the meaning of messages can be extrapolated by mapping them to the corresponding target objects (or some of their properties) \citep[e.g.][]{choi2018multiagent,ohmer-etal-2022-emergence}.
While training algorithms for more elaborate MARL setups are readily available, for example MAPPO and G2A \cite{yu2022surprising, liu2020multi}, measuring when communication emerges and is useful in such setups and analyzing the agents' communication strategy is not trivial. 
In this study, we show how existing interpretability methods, such as \emph{saliency maps} \cite{simonyan2013deep}, \emph{perturbations} \cite{greydanus2018visualizing}, and \emph{diagnostic classifiers} \cite{hupkes2018visualisation}, can be used to address these difficulties.

In sum, we make the following contributions:
\begin{enumerate}
    \item We introduce two new open-ended reinforcement learning environments for studying the emergence of language.
    \item We demonstrate that the emergent language in these open-ended settings exhibits traits that cannot emerge in classical reference games, such as sparse language use and communication of spatial locations.
    \item We show how interpretability methods can be used as general tools for tracking the emergence of language and interpreting the developed protocols. 
\end{enumerate}

Our study not only advances our understanding of how MARL frameworks can foster language emergence but also sets the groundwork for future explorations into the dynamic and adaptive nature of multi-agent communication systems.

\section{Related Work}

\begin{figure*}[t]
     \centering
     \begin{subfigure}[b]{0.49\textwidth}
         \centering
         \includegraphics[width=0.8\textwidth]{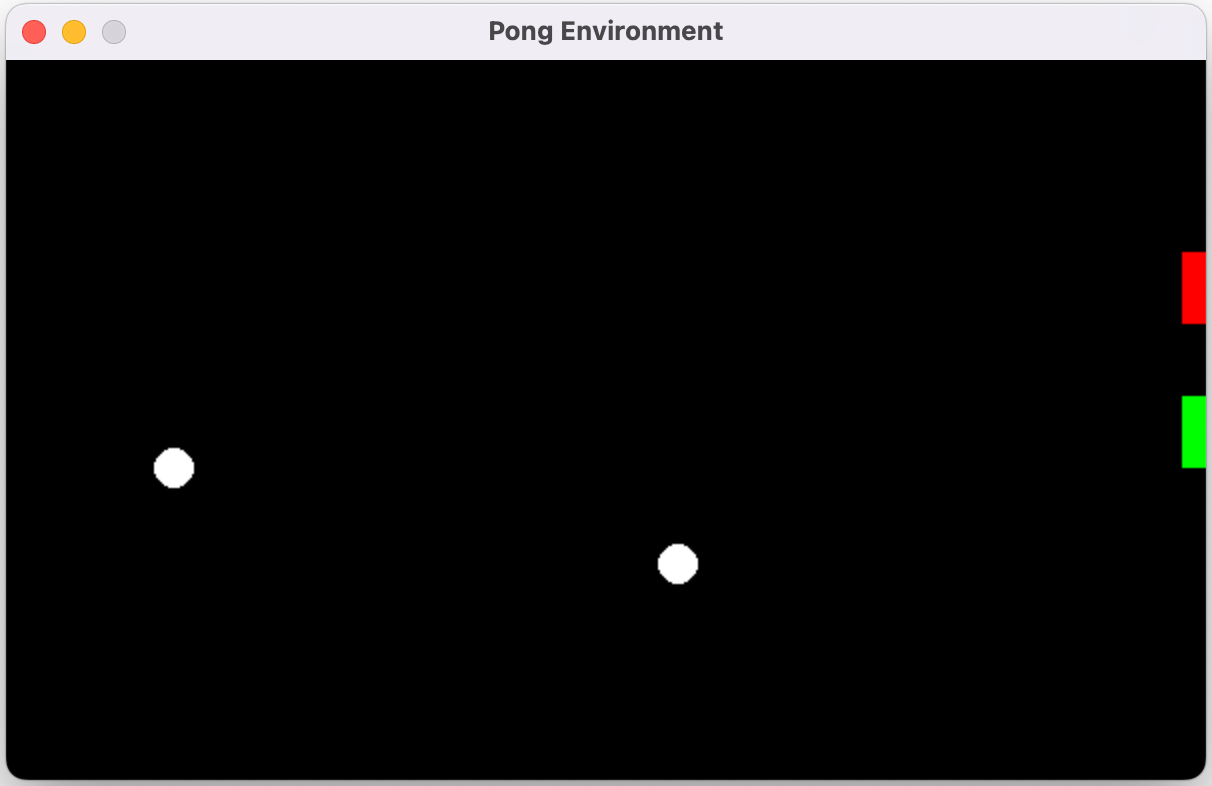}
         \caption{\centering The Pong environment with 2 players (red and green) on the right and 2 balls (white).}
         \label{PongScreenshot1}
     \end{subfigure}
     \begin{subfigure}[b]{0.49\textwidth}
         \centering
         \includegraphics[width=0.8\textwidth]{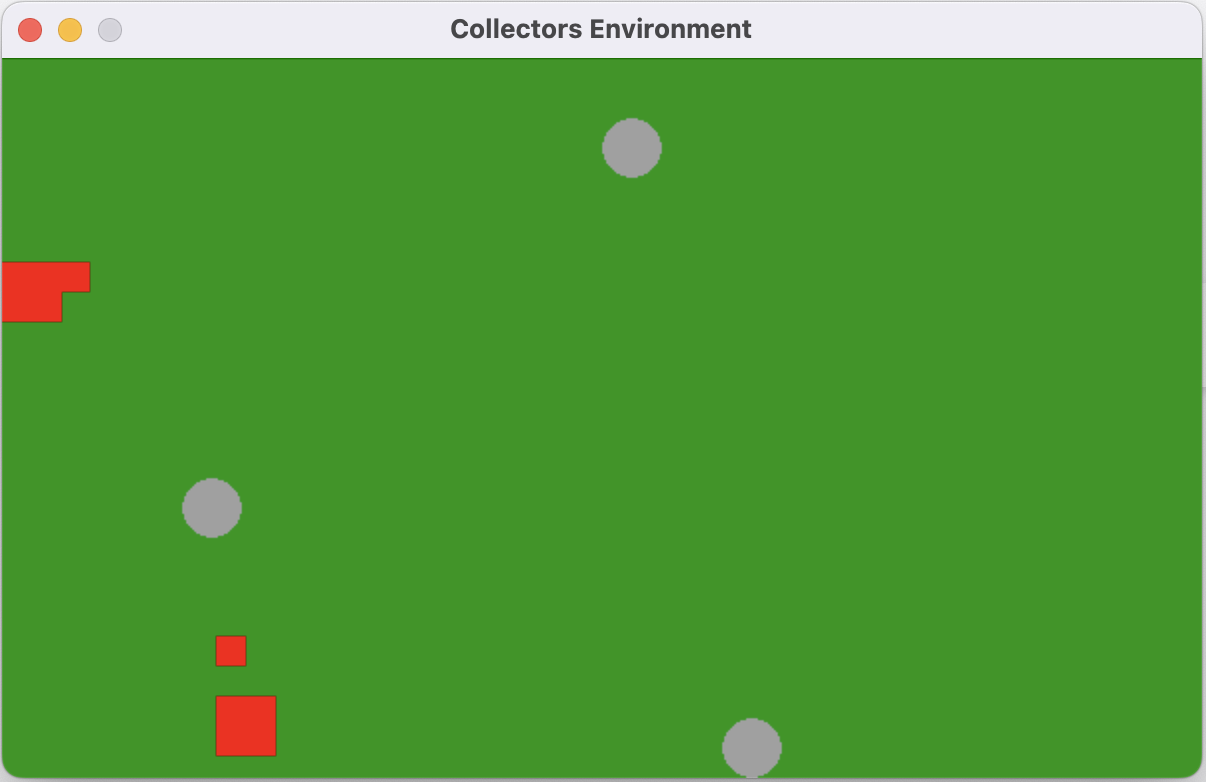}
         \caption{\centering The Collectors environment with 2 players (red), their direction of movement (smaller red dots) and 3 targets (grey).}
         \label{CollectorScreenshot1}
     \end{subfigure}
     \caption{The two environments used for our experiments.}
\end{figure*}

Our work builds a bridge between existing research on LE, especially approaches studying LE with DNN agents in increasingly realistic scenarios, and more advanced machine learning approaches to multi-agent coordination. In addition, we show how machine learning interpretability methods can be used to track the emergence of language and analyze the agents' communication strategies.

\subsection{Language Emergence Simulations}\label{related_work:LE_with_DNNs}

Research on LE has a long interdisciplinary tradition. 
While evolutionary linguistics explores the origins and evolution of human and animal communication \cite{cangelosi_2002_simulating, kirby_2002_overview, wagner_2003}, AI research aims to develop artificial agents capable of flexible and goal-directed language use, grounded in interaction \cite{steels_2003, lazaridou_2020}.

As mentioned above, the work of \citet{foerster_2016} and \citet{lazaridou_2017} started a trend towards LE setups with DNN agents, most of which employ simple reference-game environments. 
Messages in these setups are usually chosen to be discrete in order to encourage communication of conceptual information rather than low-level features as well as to facilitate an interface with natural language, for example to analyze linguistic structure \cite[e.g.][]{Ren2020Compositional,chaabouni-etal-2020-compositionality,van-der-wal-etal-2020-grammar,ohmer-etal-2022-emergence}. 
The agents are typically trained with simple RL algorithms, such as REINFORCE \cite{williams1992simple}.


To address limitations of the reference game, the field has begun to move towards more realistic scenarios. 
For instance, \citet{cao2018emergent} employed a semi-cooperative communication game with multi-turn interactions where agents negotiate over resources; \citet{harding-graesser-etal-2019-emergent} studied contact linguistic phenomena using populations of agents; and \citet{liang_2020} examined the impact of competitive pressures in a mixed cooperative-competitive setting. 
While \citet{chaabouni2022emergent} did use a reference game, they increased simulation complexity across multiple dimensions -- scaling the number of agents, possible objects, and distractors -- and employed more complex RL training techniques.  
A notable example involving \emph{situated} agents is provided by \citet{mordatch_2018}, who explored the emergence of both compositional and nonverbal communication through a simulation of multiple agents acting in a continuous 2D environment.
We draw inspiration from these established MARL environments to study LE in complex tasks with situated agents, focusing on bidirectional communication.

\subsection{(Emergent) Communication in MARL}\label{related_work:multi-agent-coordination}

MARL addresses problems involving multiple agents that are distributed in a shared environment, such as autonomous driving and robotics. The agents typically operate under partial observability in a non-stationary environment and employ RL techniques to develop cooperative, competitive, or mixed behaviors. Communication among agents may enhance their coordination and learning stability, and the MARL community has also explored \emph{emerging} communication as a more adaptive alternative to pre-specified protocols~\cite[for a survey, see][]{zhu_2024}.

Seminal work in this area includes RIAL and DIAL \cite{foerster_2016}, as well as CommNet \cite{sukhbaatar_2016}, all developed for cooperative settings using centralized training. RIAL uses discrete communication channels and Q-learning, whereas DIAL uses continuous channels and exploits the resulting differentiability for end-to-end backpropagation across agents. Later work explored emergent communication in more complex multi-agent coordination problems. For example, IC3Net transitioned from global to individualized rewards to improve training efficiency, scalability, and credit assignment~\cite{singh2018individualized}. It was successfully applied to semi-cooperative and competitive games, and a message gating mechanism allowed agents to learn \textit{when} to communicate based on scenario and profitability.

\subsection{Interpretability Methods}\label{related_work:interpretability}

Given that deep learning models have a large number of parameters and do not require any prior feature engineering, explaining their decisions is challenging. In response, explainable/interpretable AI research has come up with various methods to analyze the decision making processes of DNNs \cite[for a survey, see e.g.][]{zhang2021survey}. While a review is beyond the scope of this paper, we provide some context on the specific methods that we use: \emph{saliency maps} to study gradient attribution during training, and  \emph{perturbations} as well as \emph{diagnostic classifiers} for post-hoc interpretation.

\textbf{Saliency maps} are a tool to calculate the importance of features for a model's decision. Saliency maps are generated by calculating the gradient of a model's output with respect to its inputs and were originally developed to study which pixels in an input image contributed most strongly to the decision of image classifiers \cite{simonyan2013deep}. In reinforcement learning the method has mainly been used for exploration \citep{atrey2019exploratory}. Although computationally efficient, this approach can be noisy and may highlight irrelevant areas, reducing their reliability \citep{kim2019saliency}. 
To address these issues, methods like integrated gradients \cite{sundararajan2017axiomatic} and smoothed gradients \cite{smilkov2017smoothgrad}  were developed, which incorporate additional steps to average out noise or random variations.
Even though they are computationally more expensive, these methods provide a more stable and meaningful measure of feature importance and have been used in various applications.

\textbf{Perturbation analyses} involve manipulating input features to see which changes cause the biggest shift in the a model's output. Significant changes in the output indicate important features \citep{samek2017explainable}. This method has gained popularity in supervised learning, especially for visual data. In reinforcement learning, perturbation has been used to increasing model robustness \cite{zhang2020robust, wang2020reinforcement, liu2022robustness}, but also as an interpretability tool \cite{greydanus2018visualizing}.

Both saliency maps and perturbations can help establish a relation between an agent's inputs and its outputs but they are only of limited use when it comes to interpreting an emerging communication protocol. 

To this end, we employ \textbf{diagnostic classifiers}, originally devised to decode information in the hidden states of recurrent neural networks \citep[][]{hupkes2018visualisation}. 
These classifiers are trained on specific hypotheses about the information encoded by the network at each time step, allowing researchers to validate and refine their understanding of the network's information processing.
We apply them to the messages sent by the agents in our environments.

\section{Methods}
This section introduces our environments, agent architecture, and optimization method.\footnote{All code is available in our code submission.
}

\subsection{Environments}
In our experiments, agents are tasked to solve environments where optimal performance necessitates the use of language-based communication. We develop two such environments: 
\begin{enumerate}

\item The \emph{Multi-Agent Pong} environment (see Figure \ref{PongScreenshot1}) involves two agents and two balls that are moving simultaneously. An agent's observation does not include the position of the other agent. Consequently, the agents must coordinate their positions to consistently catch both balls. When an agent successfully catches a ball, it receives a reward of +1; if the agents miss a ball, they both receive a reward of -1 and the episode terminates.

\item In the \emph{Collectors} environment (see Figure \ref{CollectorScreenshot1}) two agents must collect a number of targets by colliding with them, while remaining unable to see each other. In contrast to Pong, agents cannot only move vertically but also horizontally. Each target has a countdown visible to the agents, within which they must be collected. If an agent successfully collects a target, it receives a reward of +1. However, if the agents fail to reach a target before its countdown expires, both agents receive a penalty of -1 and the episode terminates. Due to the distance and spawn frequency of the targets, agents cannot consistently achieve this task without utilizing the language channel for coordination.
\end{enumerate}

\subsection{Agent Architecture}
In our experiments, both agents share a network and engage in self-play to enhance coordination and learning. Each agent's architecture consists of separate actor and critic networks, both employing a three-layer dense network structure. 
We use a centralized critic that can observe both agents simultaneously, making it easier to assess their level of collaboration and, consequently, to estimate a more accurate value function.
While popular in existing literature \cite[e.g.,][]{mordatch_2018, dagan2020co}, we do not use any form of recurrence, simplifying the analysis of the language's impact on actions and resulting in faster training times.

The agents receive symbolic inputs from the environments and communicate through a language channel, where each message \( M \) is characterized by sequence length \( L \) and vocabulary size \( |V| \) with discrete tokens. The language input is handed over to the agents in a one-hot encoded format. At each time step \( t \), an agent receives the message \( m_{t-1} \) generated by the other agent in the previous time step.

\subsection{Optimization}
We use Proximal Policy Optimization (PPO) \cite{schulman2017proximal} in a self-play configuration for training our agents in the described bidirectional communication game, as PPO has previously been shown to work well in multi-agent setups without additional adaptions \cite{yu2022surprising}. Our implementation is based on the one provided by \citet{huang2022cleanrl}. PPO directly uses the game rewards to update the agents' behavior, supplemented by entropy regularization in the loss function to encourage exploration \cite{pmlr-v48-mniha16}. For optimization, we use the Adam optimizer \cite{kingma2014adam} with an initial learning rate of \( 1 \times 10^{-4} \), which we linearly decrease throughout training to a target value determined by \( \text{lr}_N = \text{lr}_0 \times \left(1 - \left(\nicefrac{N-1}{N}\right)\right) \), where the initial learning rate is \( \text{lr}_0 \) and the total number of updates is \( N \). More details regarding our hyperparameters can be found in Appendix A.

\section{Experiments and Analyses}

In the following, we detail our experiments and analyses. 

\subsection{Baseline Comparison}

To determine whether language emerges in our setup, we begin with a baseline comparison. This involves training a model to solve the environment without utilizing an active language channel. The performance of this model serves as a reference point, allowing us to assess the impact of incorporating token-based communication on task efficiency and overall success in the following training runs that include a language channel of various sizes.

We denote the vocabulary size by $|$\textbf{V}$|$, the sequence length by \textbf{L}, and the total amount of time steps during training by \textbf{TS}.

\begin{table}[h!]
\centering
\begin{tabular}{|l|c|c|c|}
\hline
\textbf{Environment} & $|$\textbf{V}$|$ & \textbf{L}                    & \textbf{TS} \\ \hline
Multi-Agent Pong & 3 & 0, 1, 2, 3 & 6e8 \\ \hline
Collectors & 4 & 0, 1, 2, 3, 4 & 1.5e9 \\ \hline
\end{tabular}
\caption{Hyperparameters used for our experiments. Sequence length $L=0$ corresponds to the baseline.}
\end{table}

\subsection{Saliency Tracking}
To understand the importance of the language channel for the agents' actions during training, we continuously calculate the current saliency values \cite{simonyan2013deep}. Saliency maps are a computationally cheap method to calculate which input features are most influential to a model's predictions by computing the gradient of its output with respect to each input feature. The saliency of a feature \( i \) is given by
\[
S_i(x) = \left| \frac{\partial F(x)}{\partial x_i} \right|
\:,
\]
where \( F(x) \) represents the model's prediction and \( x_i \) denotes the value of the \( i \)-th input feature. 

To track the emergence of language, we repeatedly calculate the gradient-based saliencies during training -- equally spaced across the training process. 
Specifically, the relevance of the agents' messages is approximated by the saliency of the received input messages with respect to the agents actions (excluding the language channel outputs), which allows us to test if a policy involves language use. 
At each of these tests, we select 10000 consecutive time steps and normalize the saliency values for each time step between 0 and 1. Message importance is quantified as the proportion of time steps where at least one saliency value exceeds a threshold of 0.8.\footnote{The choice of threshold value does not qualitatively affect our results.}

\begin{figure*}[t]
     \centering
     \begin{subfigure}[b]{0.49\textwidth}
         \centering
         \includegraphics[width=\textwidth]{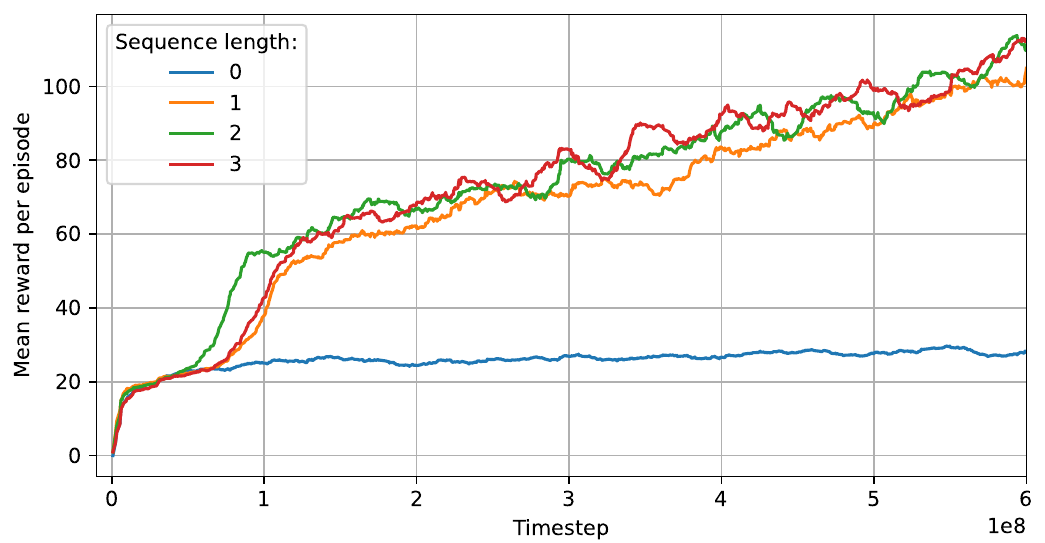}
         \caption{Training performance in the Pong Environment}
         \label{fig:PongTraining}
     \end{subfigure}
     \begin{subfigure}[b]{0.49\textwidth}
         \centering
         \includegraphics[width=\textwidth]{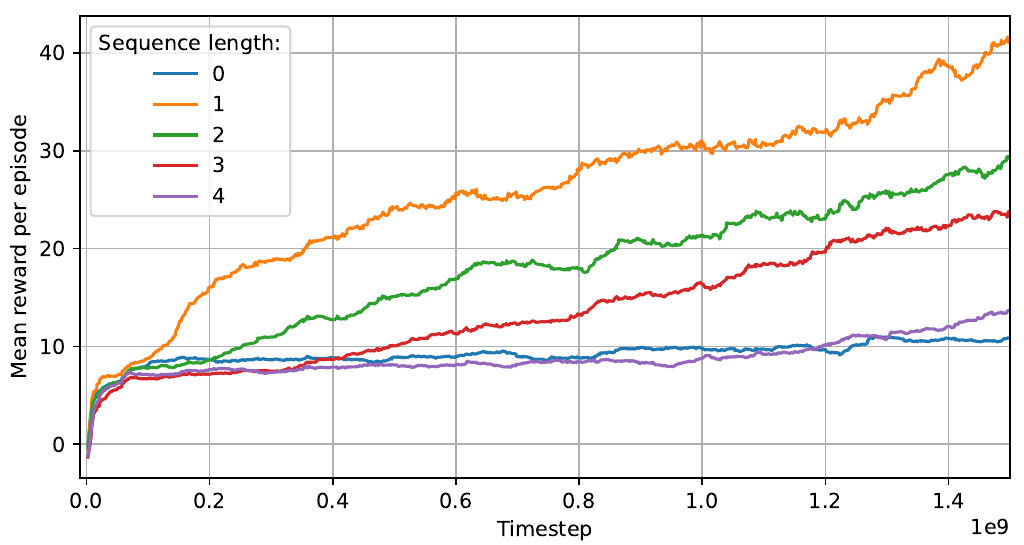}
         \caption{Training performance in the Collectors Environment}
         \label{fig:CollectorTraining}
     \end{subfigure}
\caption{Average length per epoch as a measure of how successful agents are at solving the environment.}
\end{figure*}

\subsection{Sensitivity Analysis}

To get a more detailed understanding of when the language channel is actually used in the decision-making process, we rely on perturbation to test the sensitivity of the agents' output to variations in the input messages. Specifically, we compute the Kullback-Leibler (KL) divergence between the original model outputs, \(\mathbf{P}_{\text{original}}\), and the outputs generated when replacing the input message with all other possible messages. The input data, \(\mathbf{X}\), is first separated into environment inputs, \(\mathbf{X}_{\text{env}}\), and language inputs \(\mathbf{X}_{\text{lang}}\). \(\mathbf{X}_{\text{lang}}\) is then replaced with sequences of one-hot encoded tokens, \(\mathbf{U}_{\text{token}}\), representing all possible vocabulary items. For each token \(t\) in the vocabulary, a perturbed input \(\mathbf{X}_{\text{perturbed}}\) is created by concatenating \(\mathbf{X}_{\text{env}}\) with \(\mathbf{U}_{\text{token}}\) along the feature dimension. 
We then generate the model's outputs ($\mathbf{P}_{\text{perturbed}}$) for these perturbed inputs.
Our final score is determined by the maximal KL divergence between the model's output distributions to the original and the perturbed inputs across all perturbations:
\[
   \max_{t} D_{KL}(\mathbf{P}_{\text{original}} \parallel \mathbf{P}_{\text{perturbed}}^{(t)})\:,
\]
where \( t \) indexes the tokens in the vocabulary. We use this value to quantify the model's sensitivity to changes in specific input tokens, highlighting the token that causes the strongest deviation in the model's output distribution.

\subsection{Noise Analysis}
To evaluate whether the perturbation experiment accurately captured the language channel's relevance across episodes, we replaced its contents with noise during specific time steps. This was done for periods where the perturbation test indicated high sensitivity to language changes, then repeated for low sensitivity and all time steps. Comparing success rates across these conditions reveals the importance of communication. If performance with noise during low sensitivity matches the no-noise condition, it suggests communication was not crucial during those steps. In contrast, a similar performance drop with noise during high sensitivity as with noise during all time steps indicates that perturbation worked for all situations in which language was relevant to successfully solving the environment.

\subsection{Language Analysis}

To understand what kind of information is communicated by the agents, we train diagnostic classifiers on specific subsets of episodes, to see whether our hypothesis of the language channel's contents are accurate.
First, we record \( N \) number of time steps, capturing observations and actions while agents operate, and apply our perturbation method to all time steps. Next, we identify the time steps where the perturbation test shows a higher sensitivity (KL divergence) to changes in the language channel than threshold \( T \). After training the classifier, its accuracy is tested against a separately created validation dataset with \( TS \) number of time steps. In our experiments, \( N = 300000 \), \( T = 0.02\), and \( TS = 30000\). For the classifier hyperparameters, see Appendix B.

We then continue from a specific hypothesis about the language content -- for example agent or target positions -- to label the recorded time steps with the corresponding labels.
Based on the generated dataset, we train one classifier to predict these labels from the language channel contents of both agents and a second classifier to predict the label from the observations including the language channel from one of the agents. This allows us to test if the information is completely contained inside the language channel or if the meaning of the token is dependent on the other observations.

\section{Results}

\begin{figure*}[t]
    \centering
    \includegraphics[width=\textwidth]{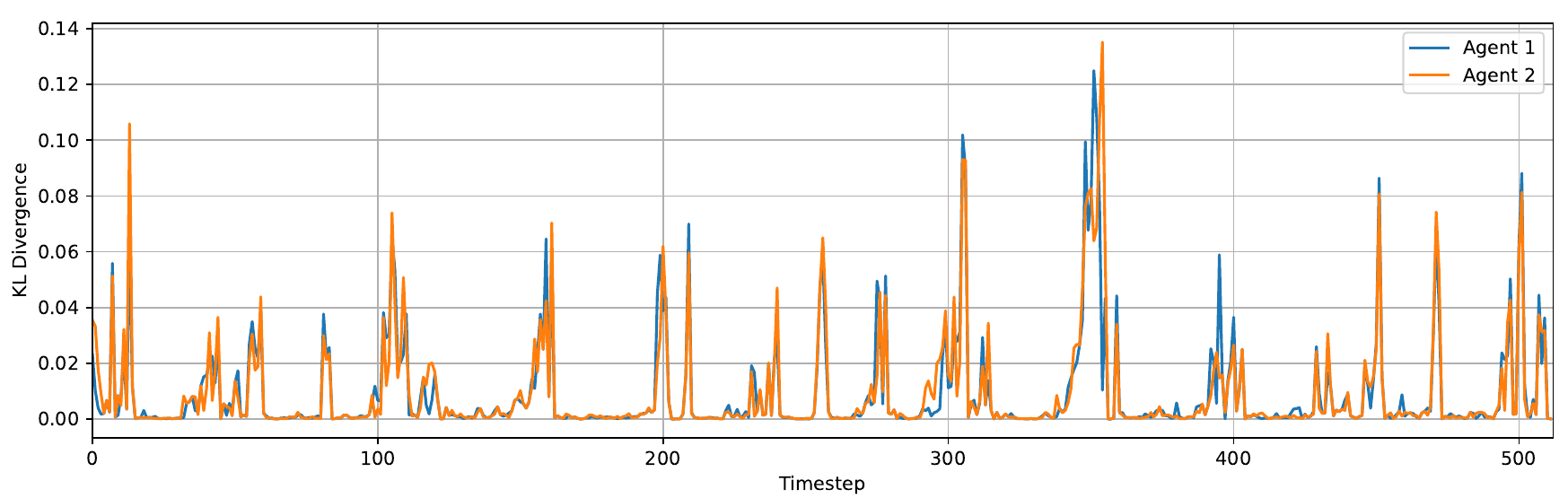}
    \caption{Sensitivity of the language channel of the agents during an episode of Multi-Agent-Pong with a sequence length of 1.}
    \label{fig:saliencies_epoch}
\end{figure*}

\subsection{Performance}

In the Multi-Player Pong environment, in Figure \ref{fig:PongTraining} we compare agents with sequence lengths of 1, 2, and 3, against the baseline without language channel.
While agents without language channel converge at a reward of roughly 25, agents with language channel continue to increase their performance to over 100. Interestingly, there is no performance difference between the different sequence lengths.

Similarly, the results for the Collectors Environment in Figure \ref{fig:CollectorTraining} show that agents equipped with a language channel significantly outperform those without. Over the entire training time, all models with a language channel beat the baseline of an average reward of 10, reaching up to an average reward of 40. We find that in this environment, an increase in sequence length lead to a decrease in learning speed, while still all models were able to beat the baseline. 
Thus, having a larger language channel is not always advantageous, as the resulting increase in action space can slow down learning. 

\subsection{Tracking the Emergence of Language}
\begin{figure}[b!]
    \centering
    \includegraphics[width=0.43\textwidth]{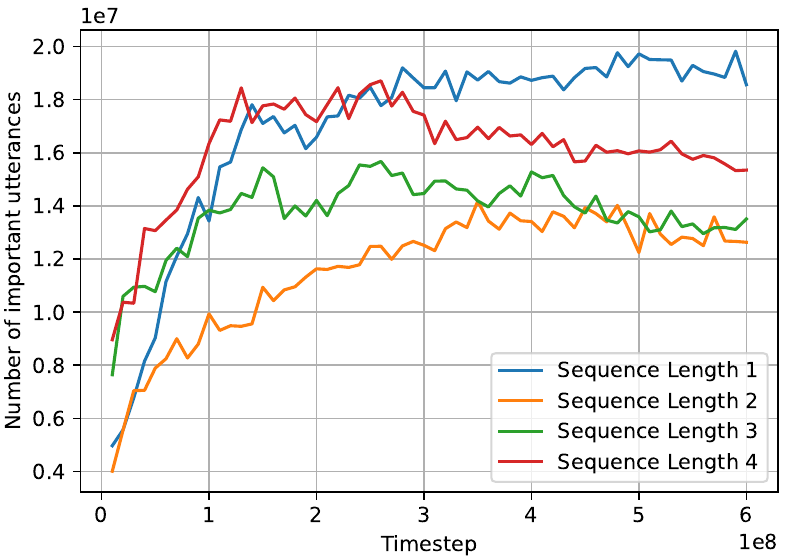}
    \caption{The number of important messages -- defined by the saliency values -- strongly influences the agents' success during training for different sequence lengths.}
    \label{fig:saliencies_live}
\end{figure}

To ensure that the advantage of agents with language channel compared to the baseline in fact arises from language-based communication and not some other difference in strategy, we evaluate the saliency scores of the language channel during training. 
The number of important utterances as measured by these saliency values (see Figure \ref{fig:saliencies_live} for Collectors and Appendix C for the Multi-Agent Pong) shows that language channel importance increases over time and closely aligns with performance gains.
However, the pattern is less clear for longer sequence lengths. 
We attribute this mainly to two factors: As we always take the maximum saliency of the entire language channel, there is a higher probability that one of the language channels has a stronger influence at the beginning, merely a) because of the random initialization of the network or b) because saliency maps are known to be sensitive to noise.

\subsection{Causal Analyses of Language Use}
To analyze the importance of language during an episode, we use the perturbation tests described above. Figure \ref{fig:saliencies_epoch} shows the importance of the language channel for one episode of the Multi-Agent Pong environment, while the same figure for the Collectors environment can be found in Appendix D. The spiking patterns indicate that the importance of the language channel fluctuates throughout the episode, suggesting that its use depends on other observations and the need for coordination at specific time steps. In the Collectors environment, we find a similar behavior, though communication here occurs more frequently. These findings are particularly significant given the challenges in traditional language emergence frameworks, where considerable effort is required to encourage sparse communication or small vocabularies while maintaining task performance, in order to foster more complex linguistic patterns \cite[e.g.][]{mordatch_2018}.

These results are further supported by our noise analysis (Table \ref{table:ma_pong_noise}, \ref{table:collectors_noise}). 
Performance drops drastically when the language channel values are replaced with noise in situations where the perturbation test indicates a high language channel importance (KL divergence $>$\emph{T=0.02}). For example, in the Collectors environment with a sequence length of 1, the average episode length decreased from 272.7 to 38.2, so below the episode length of a model without a language channel. A similar decline can be observed in the Pong environment. Conversely, performance remains nearly stable when the language channel is replaced with noise when the perturbation test indicates low language channel importance (KL divergence $>$\emph{T=0.02}). Importantly, performance never drops to chance level, even when the language channel is replaced with noise at all times (\emph{All Noise}), supporting our finding that the agents only make their actions dependent on the input messages when necessary.

\begin{table}[h]
\centering
\begin{tabular}{|c|c|c|c|c|}
\hline
\textbf{Seq} & \textbf{No Noise} & \textbf{\textless T=0.02} & \textbf{\textgreater T=0.02} & \textbf{All Noise} \\
\hline
1 & 844.0 & 803.1 & 443.9 & 460.4 \\
\hline
2 & 814.9 & 809.6 & 438.8 & 427.4 \\
\hline
3 & 876.3 & 845.9 & 379.3 & 436.7 \\
\hline
\end{tabular}
\caption{Average episode length in the MA Pong Noise Test}
\label{table:ma_pong_noise}
\end{table}

\begin{table}[h]
\centering
\begin{tabular}{|c|c|c|c|c|}
\hline
\textbf{Seq} & \textbf{No Noise} & \textbf{\textless T=0.02} & \textbf{\textgreater T=0.02} & \textbf{All Noise} \\
\hline
1 & 231.1 & 256.4 & 47.0 & 47.0 \\
\hline
2 & 187.0 & 180.5 & 55.0 & 52.8 \\
\hline
3 & 144.7 & 150.6 & 58.1 & 57.7 \\
\hline
4 & 99.4 & 84.9 & 83.4 & 76.2 \\
\hline
\end{tabular}
\caption{Average episode length in the Collectors Noise Test}
\label{table:collectors_noise}
\end{table}

\subsection{Interpreting the Messages}
Having learned that the agents successfully use the language channel, we aim to decode the information content of their messages. 
For each environment, we test one hypothesis about this content using a diagnostic classifier.
\begin{enumerate}
\item For the \emph{Multi-Agent Pong environment}, we hypothesize that the agents talk about their positions relative to each other in order to understand which ball each of them has to catch. We create binary labels for each time step indicating which agent is higher (along the y-axis)

\item For the Collector environment, we assume that the agents are talking about who should pick which target. Again, classifier inputs are defined by the language channel values, and labels are generated by encoding the target index (max 3) each agent has moved towards five time steps after sending these messages.
\end{enumerate}

Table \ref{table:diagnostic_classifiers} shows the accuracy of the two types of classifiers for each environment.
Our hypothesis for the Multi-Agent Pong environment holds with approximately 70\% accuracy when only trained on the language channel (\emph{Lang}) and 90\% when trained on the entire observations (\emph{Obs}) of one agent. In the Collector environment, our hypothesis appears to be more accurate, as the classifier reaches close to 80\% accuracy when only trained on the language channel and 95\% percent when trained on the entire observations. The observed decline in accuracy for higher sequence lengths coincides with the reduced performance during the training process.

It seems that models trained on the complete observations of one agent yields higher accuracy. However, interpreting these results is complex as observational data alone may contain (indirect) information about the target variable. 
For instance, in the game of Pong, if one agent is positioned near a target, the classifier could approximate the position of the other agent as close to the alternative target. Thus, higher accuracies on all observations might either indicate that language and environment features together provide the hypothesized information, or that environment features alone can in some cases contribute that information.
The fact that we train on time steps where language-based coordination is crucial, makes the latter explanation less likely: language is important precisely because the environment information is ambiguous. 

\begin{table}[h!]
\centering
\begin{tabular}{|c|c|c|c|c|}
\hline
\multirow{2}{*}{\textbf{Seq}} & \multicolumn{2}{c|}{\textbf{Multi-Pong}} & \multicolumn{2}{c|}{\textbf{Collectors}} \\ \cline{2-5} 
 & \textbf{Lang} & \textbf{Obs} & \textbf{Lang} & \textbf{Obs} \\ \hline
\textbf{1} & 0.683 & 0.882 & 0.789 & 0.955 \\ \hline
\textbf{2} & 0.724 & 0.933 & 0.779 & 0.948 \\ \hline
\textbf{3} & 0.714 & 0.931 & 0.786 & 0.944 \\ \hline
\textbf{4} &  &  & 0.655 & 0.933 \\ \hline
\end{tabular}
\caption{The accuracy of our diagnostic classifiers when trained only on the language channel (Lang) and the full observations of an agent (Obs)}
\label{table:diagnostic_classifiers}
\end{table}

\section{Discussion}
Our results demonstrate that open-ended environments like ours, where language is not always required, offer a novel way to study language phenomena that are difficult to explore with traditional setups such as reference games.
To analyze LE in these environments, we introduce a novel toolbox inspired by interpretability research: LE can be tracked using saliency maps, critical moments for language use can be pinpointed through perturbation analysis, and message content can be examined with diagnostic classifiers.
A key finding is that the agents resort to  using the language channel only when it is necessary to solve the environment. In fact, when the current state doesn't require coordination, the impact of language channel on the actions is so minimal that replacing the language channel with random noise does not affect the agents' ability to solve the task. This finding is surprising, considering that there is no additional cost for the agents to use language and that classic reference games often require additional loss functions for the agents to minimize the use of the language channel.

Looking ahead, our current approach could be extended by incorporating recurrency into the model architecture. At present, our model generates actions based solely on current observations, which made analyzing the emergent behavior significantly easier as we do not have to consider dependencies over multiple time steps. Introducing recurrency would enable the exploration of more complex tasks that involve dependencies across multiple time steps.

Such an extension would also require adaptations of our methodology. For example, perturbation analysis may need to account for the potential influence of changes in the language channel on hidden states over time, rather than just on immediate actions. This could offer a more detailed understanding of how information is processed within the agents.
Additionally, the diagnostic classifiers could be adjusted to consider the hidden states of the recurrent model, which might contain important information communicated across previous time steps. Incorporating recurrency into the model could thus expand the range of our analysis, allowing for a deeper exploration of the patterns of information flow and decision-making within the agents.
Furthermore, we use successfully use saliency maps to track the importance of the language channel during training. Saliency maps are cheap to compute and therefore allow for continuous calculation over time. However, saliency values become increasingly more unreliable with larger sequence lengths.
In line with \citet{atrey2019exploratory}, we thus recommend that saliency should be employed primarily as an exploratory tool.
More robust methods, such as integrated gradients, may be used for more detailed analyses -- at the cost of higher compute requirements.

\section{Conclusion}

Classic reference game setups, while prevalent in language emergence simulations, often fail to capture several critical aspects of communication. To address this, we introduced exemplary environments where situated agents can continuously interact and communicate. 
Furthermore, we demonstrated how methods from interpretability research can be adapted to overcome the challenge of analyzing emerging communication in these more complex environments.
Our analyses revealed that key features, such as communicating only when it is relevant, naturally emerge in these scenarios. 
With this, we aim to pave the way for more complex emergent communication simulations.


\newpage

\section*{Acknowledgements}
This research was conducted within the scope of MicrocosmAI, which made this work possible. For further information and access to the code, please visit microcosm.ai. The work was funded by the Deutsche Forschungsgemeinschaft (DFG, German Research Foundation) – 321892712.

\bibliography{custom,anthology}

\newpage

\appendix


\subsection{Appendix A: Hyperparameters PPO}

\begin{table}[h!]
\centering
\begin{tabular}{|c|c|c|}
\hline
\textbf{Hyperparameter}        & \textbf{Value/Type} \\ \hline
num-minibatches          & 256 \\ \hline
update epochs           & 4 \\ \hline
hidden layers          & 128, 64  \\ \hline
optimizer             & Adam \\ \hline
Advantage normalization             & True \\ \hline
\end{tabular}
\caption{Further PPO Hyperparameters}
\label{tab:updated_hyperparameters}
\end{table}

\subsection{Appendix B: Hyperparameters Diagnostic Classifiers}

\begin{table}[h!]
\centering
\begin{tabular}{|c|c|c|}
\hline
\textbf{Hyperparameter}        & \textbf{Value/Type} \\ \hline
Samples (training/testing) & 300,000/30,000 \\ \hline
Importance Threshold & 0.02 \\ \hline
batch-size           & 32 \\ \hline
learning rate        & 0.001 \\ \hline
epochs           & 120 \\ \hline
hidden layers          & 64, 32  \\ \hline
loss function         & CrossEntropy \\ \hline
optimizer             & Adam \\ \hline
\end{tabular}
\caption{Hyperparameters for die Diagnostic Classifiers}
\label{tab:updated_hyperparameters}
\end{table}

\subsection{Appendix C: Saliency Tracking Multi-Agent Pong}

\begin{figure}[h!]
    \centering
    \includegraphics[width=0.43\textwidth]{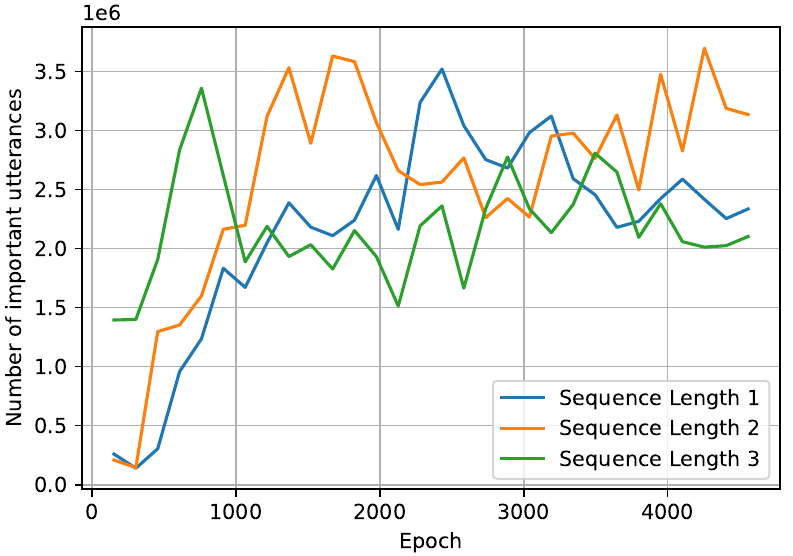}
    \caption{The number of inputs utterances strongly influencing the agents' actions during training for different sequence lengths. An utterance is important when the at least one normalized salience reaches a value of 0.8.}
\end{figure}

\newpage
\subsection{Appendix D: Perturbation test in the Collectors Environment}
\begin{figure}[h!]
    \centering
    \includegraphics[width=0.43\textwidth]{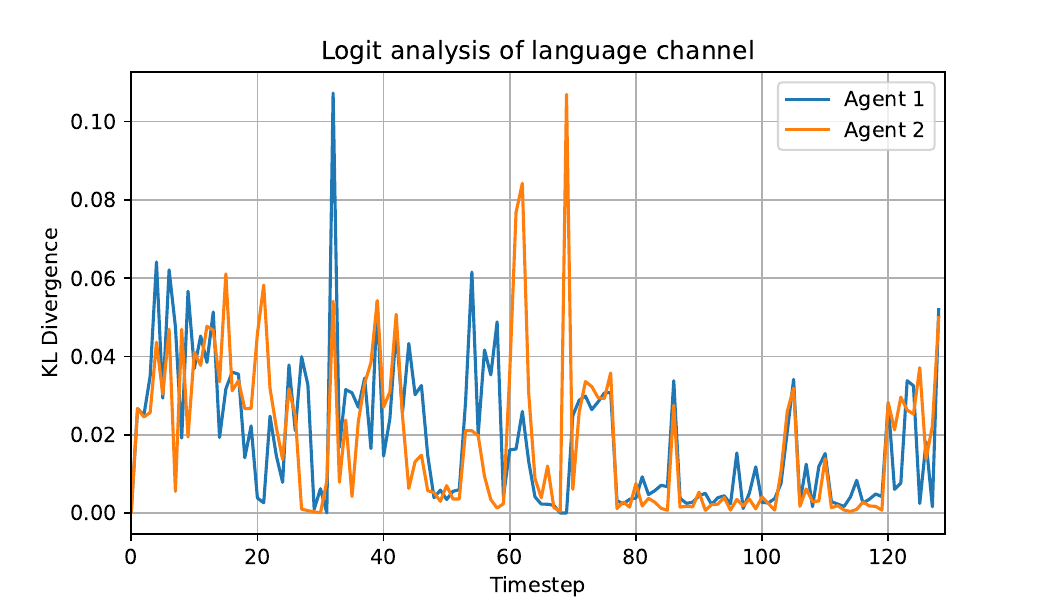}
    \caption{Sensitivity of the language channel of the agents during an episode of Multi-Pong with a sequence length of 1}
\end{figure}


\end{document}